\documentclass{article}

\usepackage{microtype}
\usepackage{graphicx}
\usepackage{subfigure}
\usepackage{booktabs} 

\usepackage{hyperref}

\usepackage{booktabs,multirow,makecell}

\usepackage{float}

\usepackage[accepted]{icml2025}

\usepackage{amsmath}
\usepackage{amssymb}
\usepackage{mathtools}
\usepackage{amsthm}
\usepackage{multirow}
\usepackage[capitalize,noabbrev]{cleveref}

\theoremstyle{plain}

\theoremstyle{definition}

\theoremstyle{remark}
\usepackage{xspace}

\usepackage[textsize=tiny]{todonotes}
\usepackage{algorithm}
\usepackage{algpseudocode}
\newcommand{\ourmethod}{AppVLM\xspace}
\newcommand{\openapp}{\texttt{open-app}\xspace}
\newcommand{\click}{\texttt{click}\xspace}
\newcommand{\longpress}{\texttt{long-press}\xspace}
\newcommand{\inputtext}{\texttt{input-text}\xspace}
\newcommand{\navigateback}{\texttt{navigate-back}\xspace}
\newcommand{\navigatehome}{\texttt{navigate-home}\xspace}
\newcommand{\wait}{\texttt{wait}\xspace}

\graphicspath{{images/}}

\icmltitlerunning{AppVLM: A Lightweight Vision Language Model for Online App Control}

\begin{document}

\twocolumn[
\icmltitle{AppVLM: A Lightweight Vision Language Model \\ for Online App Control}



\icmlsetsymbol{equal}{*}

\begin{icmlauthorlist}
\icmlauthor{Georgios Papoudakis}{equal,hw}
\icmlauthor{Thomas Coste}{equal,hw}
\icmlauthor{Zhihao Wu}{hw}
\icmlauthor{Jianye Hao}{hw}
\icmlauthor{Jun Wang}{ucl}
\icmlauthor{Kun Shao}{hw}
\end{icmlauthorlist}

\icmlaffiliation{hw}{Huawei Noah's Ark Lab}
\icmlaffiliation{ucl}{University College London}

\icmlcorrespondingauthor{Kun Shao}{shaokun2@huawei.com}


\vskip 0.3in
]



\printAffiliationsAndNotice{\icmlEqualContribution} 

\begin{abstract}

The utilisation of foundation models as smartphone assistants, termed app agents, is a critical research challenge. These agents aim to execute human instructions on smartphones by interpreting textual instructions and performing actions via the device's interface. 
While promising, current approaches face significant limitations. Methods that use large proprietary models, such as GPT-4o, are computationally expensive, while those that use smaller fine-tuned models often lack adaptability to out-of-distribution tasks.
In this work, we introduce AppVLM, a lightweight Vision-Language Model (VLM). First, we fine-tune it offline on the AndroidControl dataset. Then, we refine its policy by collecting data from the AndroidWorld environment and performing further training iterations.
Our results indicate that AppVLM achieves the highest action prediction accuracy in offline evaluation on the AndroidControl dataset, compared to all evaluated baselines, and matches GPT-4o in online task completion success rate in the AndroidWorld environment, while being up to ten times faster. This makes AppVLM a practical and efficient solution for real-world deployment.

\end{abstract}

\section{Introduction}

The development of smartphone assistants using foundation models is an open research challenge. These assistants, which we refer to as app agents, should be capable of executing human instructions on a smartphone, interacting with apps through the same interface as a human user. The user provides a textual description of a goal, and the app agent must take a sequence of actions to successfully complete the task.  Such technology has the potential to revolutionise smartphone interactions, providing significant business value by enabling automation for productivity tools, customer service, and accessibility features. Moreover, it could enhance smartphone accessibility for a wider range of users, including individuals with disabilities or those less familiar with digital interfaces.  

Two primary approaches have been explored for developing app agents. The first relies on large foundation models, such as GPT-4, combined with prompt engineering methods to solve tasks. While flexible, this approach is expensive, both in terms of financial resources and execution time; making real-world deployment impractical. 
The second approach focuses on fine-tuning smaller models  \citep[e.g.,\ ][]{digirl, ma2024comprehensive, christianos2024lightweight, wang2024distrl}, typically using an offline dataset and, in some cases, incorporating online-collected trajectories. While these methods have demonstrated promising results, many evaluations are limited to offline action predictions or online tasks drawn from the same distribution as the training dataset. However, findings from \citet{chen2024spa} suggest that when these models are tested in out-of-distribution (OOD) settings, their success rates drop significantly. This highlights a critical challenge in generalising beyond the training distribution.  

In this work, we propose \ourmethod, an app agent designed to overcome these challenges by achieving both efficiency and strong generalisation to tasks OOD compared to the original offline dataset. Our model is lightweight, enabling fast and cost-effective inference for real-time execution, and capable of adapting to OOD tasks, unlike standard offline-trained models.  
To achieve this, we assume access to an offline dataset of near-optimal human trajectories of phone interactions, which we use for Supervised Fine-Tuning (SFT) as an initial step on top of a pretrained vision-language model (VLM). This allows the model to become familiar with the observations and actions required for interacting with an Android smartphone. We then introduce a Reinforce Fine-Tuning (RFT) pipeline, consisting of data collection, utilising a distributed client-server architecture to balance resources and enable efficient data collection, followed by offline fine-tuning, where the collected data is used to refine the agent's decision-making capabilities.  
Using this pipeline, we iteratively fine-tune our model, which we refer to as \ourmethod. 

Our main contributions are summarised as follows:
\begin{itemize}
    \item We develop AppVLM, the first lightweight (3B) VLM agent capable of successfully solving tasks in the AndroidWorld environment.
    \item We outperform GPT-4o baselines and fine-tuned models on both the in-domain-data (IDD) and OOD AndroidControl test sets, achieving state-of-the-art results on this dataset to the best of our knowledge.
    \item We demonstrate that \ourmethod achieves performance comparable to GPT-4o baselines in the AndroidWorld environment, exceeding some, and only coming 4\% short of the best-performing one, while operating at a fraction of GPT-4o’s cost in both time and resources. Additionally, to the best of our knowledge, it outperforms all non-proprietary and fine-tuned models.
\end{itemize}
By striking a balance between efficiency and generalisation, \ourmethod provides a practical and scalable solution for real-world app agents, bridging the gap between foundation models and robust smartphone automation.

\section{Related Work}

\subsection{Prompt Engineering Agents}

Several recent works focus on developing agents that execute actions in smartphone or desktop environments in order to complete textual commands. With the advancement of foundation models, the research community has been exploring ways to leverage the general cross-domain knowledge of these pretrained models for app control. \citet{appagent,mobileagent} were some of the first works that utilised large foundation models to perceive smartphone observations and generate human-like actions. To successfully solve more complex tasks requiring long-term planning and history awareness, several frameworks were proposed with dedicated prompt-engineering components for steps like planning, reflection, etc. \citep{mobileagentv2, wang2024oscar, song2024mmac}. Although these added reasoning steps improved performance considerably, they significantly increased the computational cost and wall-time of each interaction. 
Other works tried to obtain app-specific knowledge utilising memory \citep{autodroid, lee2024mobilegpt}, which stores past interaction between the agent and specific apps.

\subsection{Fine-Tuned Agents}

To address the gap between the general capabilities of foundation models and the specific needs of smartphone environments, as well as to reduce the cost of querying general foundation models, several works have focused on fine-tuning to implement more specialised app agents. \citet{wang2024ponder,gou2024navigating} use large foundation models for the high-level proposal of actions or plans, while they fine-tune a smaller VLM to ground this action. \citet{ma2024comprehensive} proposed CoCoAgent, a small foundation model that aims to predict actions for app control in smartphones by decomposing the actions into action type prediction and optionally the target UI element that this action will be applied to.
Similarly, LiMAC \citep{christianos2024lightweight} introduced a small action transformer to predict the action type and the target UI element, while integrating a fine-tuned VLM for text completion. 
InfiGUIAgent \citep{liu2025infiguiagent} proposed a two-stage fine-tuning process, which first focuses on learning details about the screenshot observations, such as predicting the text of specific UI elements, and then learns how to generate actions based on user's instructions. 

Previous research has also investigated online optimisation of app agents to overcome the limitations of trajectory diversity in static datasets. DigiRL \citep{digirl} introduced an online RL framework that simulates app control tasks, training a policy that is first fine-tuned on an offline dataset. DistRL \citep{wang2024distrl} enhanced the training efficiency with asynchronous online learning. However, both methods depend on online tasks that follow the same distribution as the offline dataset. In contrast, our work aims to enable agents to tackle tasks beyond those encountered during the initial SFT within the offline dataset.

\section{Methodology}

\begin{figure*}
    \centering
    \includegraphics[width=0.9\linewidth]{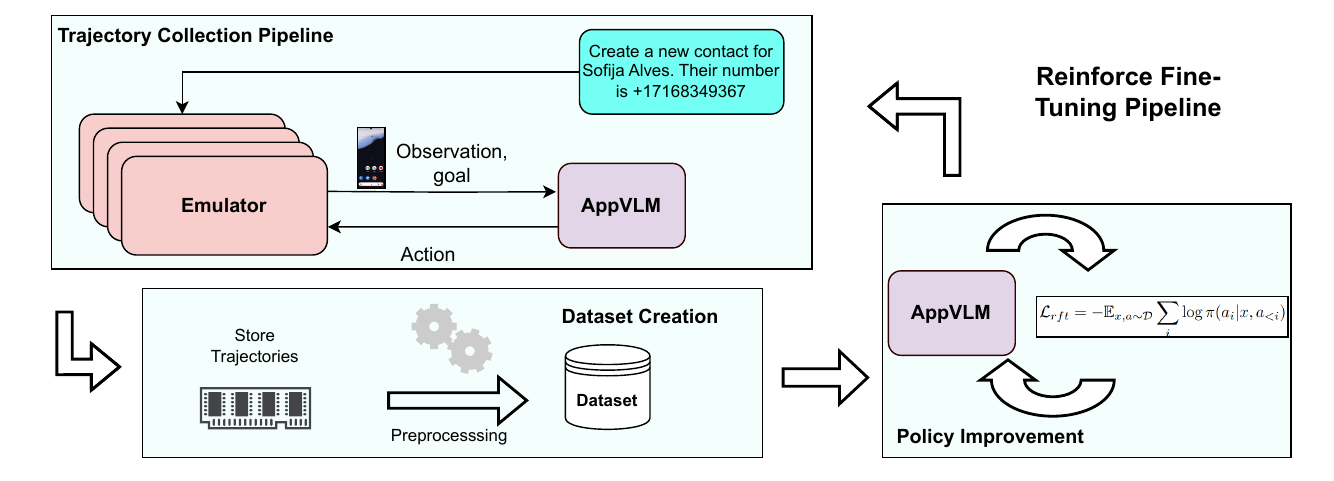}
    \caption{Visualisation of the RFT pipeline. Data is gathered by interactions between the emulators and AppVLM. The data is gathered, preprocessed and added to the dataset. It is used to perform a fine-tuning step.}
    \label{fig:appvlm}
\end{figure*}

\subsection{Problem Formulation}

We define the app control task as a Goal-conditioned Partially Observable Markov Decision Process (GPOMDP), represented as \( (\mathcal{S}, \mathcal{A}, \mathcal{O}, \mathcal{G}, R, T, \Omega) \). Here, \( \mathcal{S} \) is the set of states, \( \mathcal{A} \) is the set of actions, \( \mathcal{O} \) is the set of observations, and \( \mathcal{G} \) is the set of goals. The function \( T \) describes the state transition dynamics, and \( \Omega \) represents the observation probability distribution. The reward function is denoted by \( R \).
We assume an agent with a parameterized policy \( \pi_\theta \), where \( \theta \) represents the policy parameters. Our objective is to optimize the following expression:
\[
\max_{\pi_{\theta}} \mathbb{E}_{g \sim \mathcal{G}} \left[ \sum_{t=0}^{H-1} \gamma^t r_t \right],
\]
where $r_t$ is the reward at time step $t$ of the episode and $H$ is the horizon of the episode.
For simplicity, we assume \( \gamma = 1 \) in this task. The reward function returns 1 when the episode terminates successfully, and 0 otherwise.
To run our experiments, we specifically use the AndroidWorld environment \citep{androidworld}, which consists of parametrised tasks to be solved in an online fashion.
For example, the task of adding a contact might be described as ``Create a new contact for Sofija Alves. Their number is +17168349367." In this case, the parameters are the first name, surname, and phone number, allowing for a vast number of task variations. Our objective is to train an agent that can solve as many tasks as possible, using a lightweight model that can generalise effectively within the parameter space.

\subsection{Supervised Fine-Tuning}
\label{sec:sft}

Before initiating any online interactions within the AndroidWorld environment, we first perform SFT on a VLM using the AndroidControl dataset \citep{androidcontrol}, to allow the model to learn essential Android phone interactions. We use the Paligemma-3B-896 \citep{beyer2024paligemma} as our base model for several reasons. First, with 3 billion parameters, it offers a good balance of performance and efficiency, making it lightweight enough for mobile device deployment, especially when quantised to lower precision.
Furthermore, Paligemma-3B-896 downscales images from their original resolution of 2400x1080 pixels to 896x896 pixels. This preserves important visual details, such as legible text, while supporting higher accuracy in tasks that require visual comprehension. In contrast, many CLIP-based \citep{radford2021learning} vision transformers typically downscale images to $224x224$ pixels, a reduction that results in the loss of fine-grained details, making it difficult to retain important visual details and hindering task success.
Paligemma-3B-896 has been fine-tuned for computer vision tasks, and is therefore not inherently capable of executing app control commands based on textual instructions. As such, the SFT step in this work is essential for adapting the model to execute app-specific tasks within the AndroidWorld environment.

The input for Paligemma is constructed as follows: For each observation, we use a screenshot annotated with bounding boxes and a label indicating the UI element number for each clickable item. This information is available in the UI tree of the observation, which can be extracted from both the AndroidControl dataset and any Android device. In addition to the visual data, the textual input includes the specified goal and the history of actions. A more detailed explanation of the observation preprocessing can be found in \Cref{sec:androidworld_env,sec:dataset}.
To reduce computational costs during both training and inference, we avoid including the full history of observations. Instead, we only include the history of recent actions, as this provides valuable context with minimal added token complexity. Similar approaches have been explored in previous research \citep{putta2024agent}.

\subsection{Reinforce Fine-Tuning}
\label{sec:rft}

After fine-tuning the agent on the AndroidControl dataset, we deploy it within an interaction and fine-tuning pipeline using the AndroidWorld environment. We refer to this procedure as Reinforce Fine-Tuning (RFT), also known as Reinforced Self-Training  (ReST) \citep{gulcehre2023reinforced}, or iterative SFT with rejection sampling. Please note, that RFT should not be confused with the on-policy REINFORCE algorithm \citep{willams1992simple}.
Collecting data and fine-tuning in this way is an essential step to enable the agent to adapt to tasks in AndroidWorld, which differ from the training data provided by AndroidControl.
The RFT pipeline consists of two steps, executed sequentially: (1) data collection and preprocessing and (2) fine-tuning our model, AppVLM. These two steps of data collection and policy improvement using the SFT loss correspond to the "grow" and "improve" steps of the ReST algorithm \citep{gulcehre2023reinforced}. In contrast to ReST, our RFT does not use the original offline dataset during the policy improvement.

\subsubsection{Data Collection and Preprocessing}
\label{sec:data_collection}

To facilitate efficient and scalable data collection, we implement a distributed client-server architecture, illustrated in \Cref{fig:appvlm}. In this setup, the Android emulator acts as the client, while the \ourmethod agent operates as the server. Tasks, along with their associated parameters and agent settings, can be submitted to a shared ``task queue". When an emulator client becomes idle, it retrieves a task from the queue and executes it. The script running the emulator will request action generations from the \ourmethod server at every step, providing the current environment observation. These requests are maintained in an ``action request" queue, to be processed sequentially by the \ourmethod server. Generated actions are then returned and executed in the client emulator, enabling it to proceed to the next step. This distributed client-server architecture enables us to run several emulators in parallel, efficiently pulling tasks from the same queue and calling \ourmethod from different machines.

To ensure diverse data collection, each task in AndroidWorld is initially added several times to the task queue, with high sampling temperatures. Then, we identify tasks that have been solved fewer than a threshold $\tau$ number of times and repeat, focusing on these tasks. The process is repeated for multiple rounds, ensuring that more challenging tasks receive increased attention and maintaining a broad representation of all tasks. 
Each successful trajectory is stored locally and preprocessed before being added to the training dataset. However, early trajectories, particularly those collected before fine-tuning on AndroidWorld, often contain erroneous actions, even when successful. To reduce these errors, we apply a filtering step before fine-tuning, comparing consecutive screenshots and removing the earlier timestep if the observation has not changed. This prevents redundant or erroneous action sequences from being reinforced during fine-tuning. After data collection and preprocessing, the final training dataset is constructed by oversampling tasks solved fewer than $\tau$ times, matching their frequency to that of more frequently solved tasks. This guarantees that the model is not biased toward tasks that were easier to solve during data collection, helping it generalise more effectively across different types of interactions. We set $\tau = 10$, such that a diverse number of samples can be used for training, but not so high that the oversampling might become extreme.

\subsubsection{RFT Policy Improvement}

After collecting a dataset that contains successful AndroidWorld trajectories, we fine-tune \ourmethod to enhance the agent’s ability to solve a broader range of tasks. The optimisation objective follows the standard maximum likelihood objective weighted by a \emph{return} term, where successful trajectories are assigned a return of one, and unsuccessful ones receive a return of zero. In practice, trajectories that have received zero return are not included in the fine-tuning dataset. The RFT optimisation objective is:

\begin{equation}
    \mathcal{L}_{rft} = -\mathbb{E}_{x,a \sim \mathcal{D}_{on}}\sum_{i} \log \pi(a_i | x, a_{<i})
\end{equation}
where $\mathcal{D}_{on}$ is the gathered preprocessed dataset.

\subsection{The Training Architecture}

\begin{algorithm}[b]
\caption{Pseudocode of \ourmethod training pipeline }
\label{pseud:pipeline}

\begin{algorithmic}[1] 

\State \# Initial VLM Fine-Tuning
\State AppVLM-base $\gets$ SFT(VLM, $\mathcal{D}$)

\State $\mathcal{D}_{on} \gets \emptyset$, AppVLM-RFT\_0 $\gets$ AppVLM-base

\For{$i \gets 0$ to $N$}
    \State \# Collect data from AndroidWorld 
    \State $\mathcal{D}_{on}$ $\gets$ $\mathcal{D}_{on} \cup $ CollectData(AndroidWorld)
    \State \# Improve the  AppVLM Policy
    \State AppVLM-RFT\_i+1 $\gets$ RFT(AppVLM-RFT\_i, $\mathcal{D}_{on}$)
    
\EndFor

\State \# Perform a final SFT on AppVLM-base
\State AppVLM $\gets$ SFT(AppVLM-base, $\mathcal{D}_{on}$)
\State return \ourmethod
\end{algorithmic}

\end{algorithm}

Having outlined the individual training steps, this section provides a summary of the entire training pipeline, presented in \Cref{pseud:pipeline}. As described in \Cref{sec:sft}, the process begins with  SFT on the pretrained Paligemma model. The resulting model, referred to as AppVLM-base, can interact with Android emulators and generate actions in the correct format. However, it faces challenges in online interactions. To address this, the model undergoes further refinement using RFT on data generated within the AndroidWorld environment. This pipeline alternates between collecting trajectories and fine-tuning the model by maximising the likelihood of actions that led to successful task completions. The policy improvement phase of the RFT is conducted offline to allow recalibrating the dataset by removing duplicate observations and to ensure a more balanced distribution of data across tasks. 
This approach prevents over-representation of simpler tasks, which could cause the model to overfit. In our experiments, we use three iterations of RFT.
As a final step, all data collected throughout the entire RFT procedure is used to fine-tune the agent of the initial SFT stage, AppVLM-base,  using the standard maximum likelihood objective. In \Cref{sec:ablation}, we present an ablation study demonstrating that this approach achieves superior performance in both online and offline settings compared to fine-tuning directly the output of the RFT pipeline.

\section{Experiments}

\subsection{AndroidWorld Environment}
\label{sec:androidworld_env}
AndroidWorld \citep{androidworld} offers a benchmark of 116 unique tasks with randomised parametrisation, leading to an infinite number of task instances. It provides a real Android environment in which agents can attempt to solve these app control tasks, evaluated with ground-truth reward signals based on the phone's state. This facilitates and standardises the execution of actions, and allows for a realistic and fair evaluation of app agents. AndroidWorld leverages the AndroidEnv \citep{androidenv} library to allow  communicating with an emulated Android virtual device.

\textbf{Actions:} AndroidWorld uses a fixed action space, as action grounding is a standard practice in app control tasks, to translate model outputs to environment actions. The action space that is used by \ourmethod is very similar to the one used in AndroidWorld and only minor action translations are required, such as for click targets. \Cref{appendix:action} further discusses our action space and AndroidWorld conversions.

\textbf{Observations:} AndroidWorld observations consist of a phone's screenshot and an accessibility UI tree that provides details about UI elements, including their text content, type, position, and attributes (e.g., whether they are clickable). We use the accessibility tree to identify clickable elements and determine their bounding box coordinates. These coordinates are then drawn as overlays on the screenshot, with numbered labels for clarity.
Each task in the AndroidWorld environment includes a high-level textual instruction, or goal, that defines the agent's objective. This goal, along with the agent's past actions, is provided as textual input to \ourmethod alongside the annotated screenshot. Further details on how we process AndroidWorld observations and the history of actions are available in \Cref{appendix:obs}.

\textbf{Agent:} To operate within the AndroidWorld environment, the VLM must be encompassed within an agent. For a given task, at each timestep, the agent receives the current goal and observation from the environment. These are fed into the model, using the observation processing described above, to generate a new action. We then convert this action into the expected format and execute it in the environment, updating the phone state. This procedure continues until either the the task is solved, for which AndroidWorld provides a reward signal dependent on phone state, or the agent fails to solve it in the allotted number of steps.

\begin{table*}[h]
\centering
\caption{Comparison across agents of action accuracy in the four splits of the AndroidControl test set.}\label{tab:results_ac}
\begin{tabular}{ll l  cccc}
\toprule
&\multirow{2}{*}{Agent}     & \multirow{2}{*}{\makecell{Input Type}}  & \multicolumn{4}{c}{Action Accuracy $\uparrow$} \\ 
\cmidrule(lr){4-7} 
& & & IDD & Task-Unseen & Cat-Unseen & App-Unseen  \\ 
\midrule
\multirow{3}{*}{\rotatebox[origin=c]{90}{\small{GPT-4o}}}
& SeeAct         & screen + UI tree                  & 31.5 & 30.7 & 30.6 & 30.9 \\ 
& T3A            & UI tree                   & 56.1 & 55.8 & 56.5 & 54.2 \\ 
& M3A            & screen + UI tree                    & 60.8 & 59.3 & 60.8 & 60.4\\ 
\midrule
\multirow{4}{*}{\rotatebox[origin=c]{90}{\small{Fine-Tuned}}}
& Llama-3        & UI tree                   &  65.5   &  58.7    &  58.3    &  57.1 \\
& LT-all-r64*    & UI tree                  & 70.8 & 59.6 & 57.4 &  58.5 \\ 
 &\ourmethod-base                & screen + b-boxes                & \textbf{73.9} & \textbf{65.9} & \textbf{65.1} & \textbf{65.4} \\ 
&\ourmethod                   & screen + b-boxes                 & 69.0 & 62.7 & 61.9 & 62.2 \\ 
\bottomrule
\end{tabular}

\end{table*}

\subsection{AndroidControl Dataset}
\label{sec:dataset}

Before the RFT in AndroidWorld, we perform an SFT step on  AndroidControl \citep{androidcontrol}, due to its similarities with AndroidWorld. 
AndroidControl is an open-source app control dataset that contains a large number of human demonstrations for a wide range of phone navigation tasks, from setting alarms to adding items to a shopping basket.

Importantly, AndroidControl episodes present themselves similarly to AndroidWorld tasks. Each episode contains a textual goal along with a list of observations and corresponding human-selected actions. Much like AndroidWorld, observations are composed of both a screenshot of the current phone screen and an accessibility UI tree. In line with our processing of AndroidWorld observations, we use the accessibility tree information to annotate the screenshot with bounding boxes and number labels. Moreover, AndroidControl actions are again grounded to a fixed action space, which, other than very minor discrepancies discussed in \Cref{appendix:action}, is identical to AndroidWorld's. As described in \Cref{sec:sft}, the final model input is formed by combining the task goal and actions from the five previous steps in text format, and the annotated screenshot in image format.

\subsection{Evaluated Baselines}

\textbf{GPT-4o methods:} T3A and M3A \citep{androidworld} were introduced alongside AndroidWorld and are widely used as evaluation baseline. They are based on the same two-step prompting method: summarising the previous action in one step, and generating an action based on the current observation and the summary. T3A is text-only, receiving observations as a list of UI elements and descriptions based on the UI tree, while M3A also receives screenshots of the phone screen annotated with UI element bounding boxes and labels. In addition to these two agents, we include SeeAct \citep{seeact}, another popular two-step GPT-prompting method. Specifically, we use the $\text{SeeAct}_{\text{choice}}$ variant, as in \citet{androidworld}, since this has been found to be the best-performing \citep{seeact}. In this variant, GPT-4o is given the goal and screenshot and prompted to produce a high-level description of the proposed action. The next step is an action grounding step, where a multiple-choice list of UI elements is provided, along with the action proposal and details about expected action formats, and GPT-4o is tasked with producing the final action output.

\textbf{Fine-Tuned Models:} We also include smaller models, fine-tuned on the AndroidControl dataset, as evaluation baselines. First, we evaluate Llama-3 \citep{dubey2024llama} with 8B parameters. We fine-tuned Llama-3 using a similar observation format as \ourmethod, but instead of using screenshots as input, we provide a condensed textual form of the UI-tree. To reduce computational requirements, we fine-tune Llama-3 using LoRA adapters \citep{hu2021lora}.
In addition, for AndroidControl, we include the action prediction accuracy of the LT-all-r64 model as reported by \citet{androidcontrol}. LT-all-r64 model is a fine-tuned version of PALM-2S using LoRA adapters, which achieved the highest accuracy among all evaluated models in \citet{androidcontrol}. To the best of our knowledge, it has achieved the highest reported accuracy to this day in AndroidControl.
It is important to note that this comparison may not be entirely consistent. While we have made every effort to faithfully reproduce their evaluation protocol, minor differences could impact the comparison with \ourmethod. Since the LT-all-r64 model is unavailable, we do not include it in online experiments. We do however similarly report the success rate of InfiGUIAgent \citep{liu2025infiguiagent} in the AndroidWorld environment as stated in its original paper. Methods with results taken directly from their papers are marked with an asterisk (*).

\subsection{Experimental Setup}

\begin{table*}[t]
\centering
\caption{Comparison of different agents in terms of success rate in the AndroidWorld environment.}\label{tab:results_aw}
\begin{tabular}{l l c c cccc}
\toprule
& \multirow{2}{*}{Method}    & \multirow{2}{*}{Size $\downarrow$} & \multirow{2}{*}{\makecell{Average \\ Infer. Time (s)}$\downarrow$}   & \multicolumn{3}{c}{Success Rate $\uparrow$} &  \multirow{2}{*}{\makecell{Overall \\ Success Rate} $\uparrow$} \\ 
\cmidrule(lr){5-7} 
 & & & & Easy & Medium & Hard  \\ 
\midrule
\multirow{3}{*}{\rotatebox[origin=c]{90}{\small{GPT-4o}}}
& SeeAct     &   - & 15.82 &   34.2 & 15.5  & 4.2  & 22.0 \\ 
& T3A        &   - & 4.29  & \textbf{64.9} & 26.2  & \textbf{14.6} & \textbf{41.9} \\ 
& M3A        &   - & 11.42  & 60.5 & 20.2  & 8.3  & 36.6 \\ 
\midrule
\multirow{4}{*}{\rotatebox[origin=c]{90}{\small{Fine-Tuned}}}
& Llama-3    &    8B & 2.35 & 31.6 & 6.0 & 4.2 & 17.5 \\
& InfiGUIAgent* &  2B & - & 25.0 & 0.0 & 0.0 & 9.0 \\

&\ourmethod-base    & 3B & 0.91                     & 21.9   & 2.4   & 2.1   & 11.4 \\ 
&\ourmethod          & 3B & 0.91                & 57.9   & \textbf{27.4}  & 8.3   & 37.8 \\ 
\bottomrule
\end{tabular}
\vspace{-2pt}
\end{table*}

Our experiments focus on two evaluations: an offline evaluation of the action prediction accuracy in AndroidControl, and an online evaluation of the success rate in the AndroidWorld environment tasks. Details about these can be found in \Cref{sec:androidworld_env,sec:dataset} respectively. Here we discuss specifically how we conduct evaluation in these settings.

\textbf{AndroidControl:} Each timestep is a datapoint, composed of a goal, observation, and an action. Models are tasked with generating an action, which will be compared against the ground truth. Fine-tuned models are trained to provide the appropriate action format, while GPT-4o methods are provided with a large prompt detailing the format, as in \citet{androidworld}. 
A relaxed action prediction accuracy is reported for all methods, whereby a click target is considered correct as long as its bounding box is within the target element, following previous works \citep{androidcontrol}.

\textbf{AndroidWorld:} 
Online evaluation is performed, where agents are tasked with taking steps until a task is either solved or the maximum number is steps is reached. Task success is evaluated at every step using the provided reward signal, and a task is considered unsuccessful if the maximum number of steps is reached. In addition to overall success rate, we report per-difficulty success rates, using the task difficulty information provided by the benchmark. Due to the nature of our agents, action space, and evaluation process, certain tasks are omitted from the evaluation, notably verification and Q\&A tasks. Discussed further in \Cref{appendix:subset}, our final benchmark consists of 82 tasks, with a harder difficulty distribution than the full 116-task benchmark. In our AndroidWorld experiments, we perform evaluation across three different seeds, leading to different task parameters (e.g. contact name), and report the average performance across runs.

\subsection{Results and Analysis}
\label{sec:results}

\Cref{tab:results_ac} shows the action accuracy of all methods on the four splits of the AndroidControl test set. We find that \ourmethod-base, which is fine-tuned only on AndroidControl, outperforms all baselines as well as  \ourmethod. It is important to highlight that \ourmethod-base achieves the best action accuracy on this task, surpassing the previous benchmark achieved by LT-all-r64. Even \ourmethod achieves comparable accuracy to LT-all-r64 in IDD test set, and higher accuracy in OOD test splits.

The decline in action accuracy  of \ourmethod compared to AppVLM-base is expected, as the final SFT step relies solely on new data from the AndroidWorld environment. This shift reduces model accuracy in AndroidControl. A key factor in this decline is the rigidity of AndroidControl action accuracy evaluation. For example, in AndroidControl, the trajectory typically includes a \wait action after performing \openapp. In contrast, AndroidWorld introduces an automatic two-second delay between actions, eliminating the need for an explicit \wait action. During the online dataset preprocessing, these \wait actions are usually removed, as they do not affect the phone’s screenshot.
Finally, \ourmethod also outperforms Llama-3 in AndroidControl, which may indicate that for the specific task of predicting actions that match the ground truth, the image may be more informative. A similar pattern is observed when comparing M3A with T3A, providing further evidence that visual information plays a crucial role in action prediction.

\Cref{tab:results_aw} presents the online evaluation success rate of \ourmethod and related baselines in the AndroidWorld environment across three different difficulty levels. \ourmethod achieves performance comparable to M3A/T3A while requiring significantly fewer resources, both in financial cost and computation time.
Indeed, it exceeds both SeeAct and M3A's performance, while coming only 4\% short of T3A's performance. Moreover, its average inference time is a fraction of GPT-4o's, with \ourmethod being over 10 times faster than SeeAct and M3A, and almost 5 times faster than T3A.
We also emphasise that \ourmethod achieves the highest success rate in AndroidWorld among all fine-tuned models.

AppVLM-base shows strong performance in AndroidWorld, despite being fine-tuned only on AndroidControl. Llama-3, which has also been fine-tuned exclusively on AndroidControl, exhibits comparable results. This suggests that Llama-3 could serve as an alternative to Paligemma as the base model for \ourmethod. However, Paligemma remains the preferred choice, as it is almost three times smaller, enabling much faster inference.
Interestingly, we observe that higher accuracy in AndroidControl does not always lead to a higher success rate in AndroidWorld, even for models fine-tuned on the same data. For example, Llama-3 outperforms \ourmethod-base and T3A outperforms M3A in AndroidWorld, but the opposite is true in AndroidControl.

\begin{figure*}[ht]
    \centering
    \includegraphics[width=0.88\linewidth]{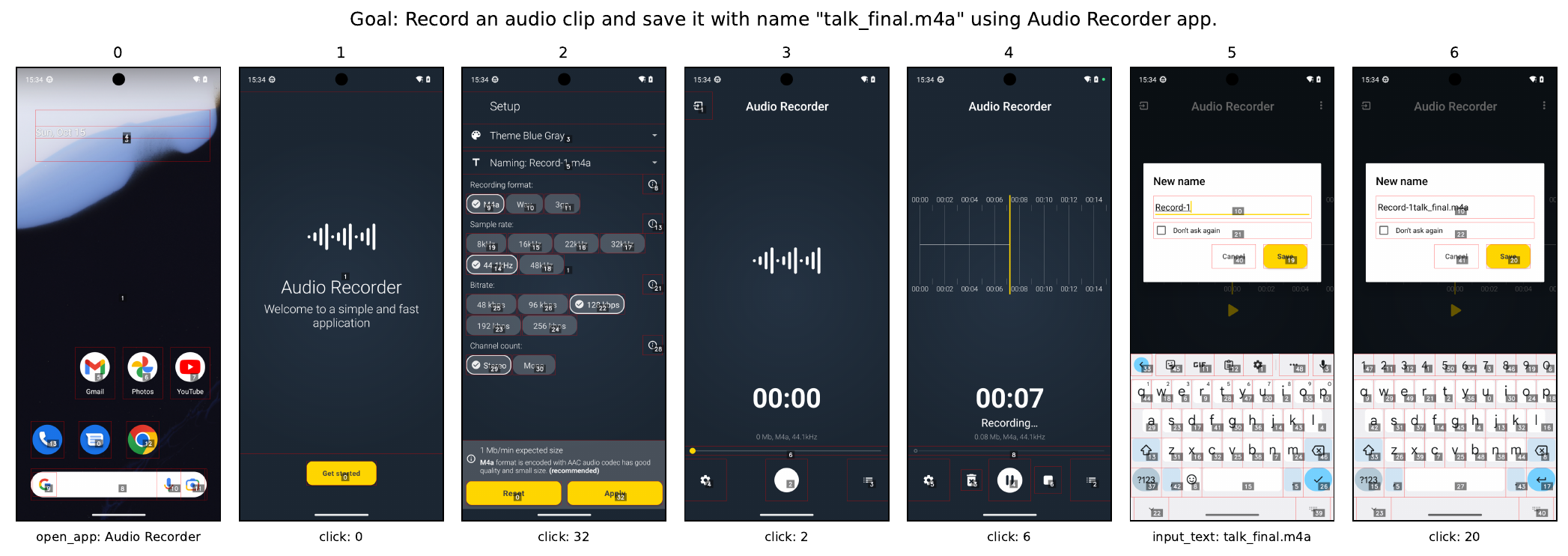}
    \caption{Example trajectory in AndroidWorld, with the goal at the top and the taken actions below each timestep's screenshot.The agent almost succeeds in solving this task, but forgets to clear the text field before typing in the penultimate step.}
    \label{fig:fail_trajectory}
\end{figure*}

\subsection{Additional Studies}
\label{sec:ablation}

\begin{table}[t]
\centering
\caption{Comparison of \ourmethod RFT iterations.}\label{tab:rlf}
\begin{tabular}{l cc }
\toprule
Agent     & AndroidControl & AndroidWorld \\
\midrule
\ourmethod-RFT\_1                & 72.5 & 17.9 \\ 
\ourmethod-RFT\_2                & 71.0 & 23.2 \\ 
\ourmethod-RFT\_3                & 66.0 & 30.5 \\ 
\ourmethod-RFT\_4                & 64.3 & 35.0 \\ 
\bottomrule
\end{tabular}
\end{table}

In this section, we provide additional studies to further explain the design choices of \ourmethod. First, in \Cref{tab:rlf}, we present the action accuracy and success rate in AndroidControl and AndroidWorld respectively for different iterations of RFT. We observe that RFT plays a crucial role in improving \ourmethod's success rate in AndroidWorld, with a linear increase over three iterations. However, further iterations of RFT showed that the improvement in success rate was lower compared to performing  SFT in AppVLM-base, as evidenced in \Cref{tab:results_ac,tab:ablation}. Additionally, the action accuracy in AndroidControl drops across RFT iterations, as previously discussed in \Cref{sec:results}.

We also provide an analysis of how the final SFT step influences both offline and online performance (see \Cref{tab:ablation}). We compare \ourmethod, against AppVLM-RFT\_4, and AppVLM-AWO (AndroidWorld Only), which skips SFT on AndroidControl entirely, performing SFT on top of the pretrained Paligemma model using the collected AndroidWorld dataset.
Our results show that AppVLM-RFT\_4 suffers from lower action prediction accuracy in AndroidControl compared to \ourmethod. This follows the downward trend observed in \Cref{tab:rlf} over successive RFT training steps. Similarly, its AndroidWorld  success rate is lower than that of \ourmethod. We hypothesise that AppVLM-RFT\_4’s performance saturates as it starts to overfit to the simpler tasks.
AppVLM-AWO, on the other hand, performs poorly in offline evaluations since it has not been fine-tuned on AndroidControl. Its online success rate is also relatively low, because it has not learned basic Android interactions that would have been acquired through SFT on AndroidControl, which also represents a much more significant amount of training data than the collected AndroidWorld dataset.
By applying the final SFT step on AppVLM-base, we retain high action prediction accuracy on AndroidControl while achieving the highest success rate on AndroidWorld tasks compared to the fine-tuning baselines.

\begin{table}[b]
\centering
\caption{Comparison of \ourmethod ablations.}\label{tab:ablation}
\begin{tabular}{l cc }
\toprule
Agent     & AndroidControl & AndroidWorld \\
\midrule
\ourmethod               & 69.0       & 37.8           \\ 
\ourmethod-RFT\_4            & 64.3         & 35.0        \\ 
\ourmethod-AWO          &  29.8         & 22.4           \\ 

\bottomrule
\end{tabular}
\end{table}

\subsection{Case Study and Failure Analysis}
\label{sec:case_study}

For illustration purposes, we show examples of  AndroidWorld trajectories in \Cref{fig:fail_trajectory} and \Cref{appendix:case_study}. \Cref{fig:fail_trajectory} demonstrates a failure case for \ourmethod. In this trajectory, the agent almost completes the task correctly, but fails to clear the existing text before adding the filename in the penultimate step. This is a common mistake, where the agent does most of the task correctly but forgets to perform one minor action.
This often happens because the AndroidControl dataset used for the initial fine-tuning does not contain tasks where such a step, for example clearing a text field, is required. In addition, because our model is relatively small, it may lack certain intuition or reasoning needed to realise it must do this. Therefore, RFT can help mitigate this behaviour if the agent learns to perform the missing action during data collection. In the next iteration, this action will be reinforced and the agent will solve the task more frequently. For example, our final model learns to delete the existing text in the task from  \Cref{fig:fail_trajectory}, as seen in \Cref{fig:success_audio_filename}. However, this type of failure still occurs in our agent for tasks that it has never managed to solve, and thus struggles to learn which step could be missing. Nevertheless, our iterative fine-tuning learns to solve many previously unsolved tasks, even taking key actions which might be absent or extremely rare in the initial fine-tuning AndroidControl dataset, such as \texttt{long-press} (see \Cref{appendix:case_study}).

\section{Conclusion}

In this work, we introduced \ourmethod, the first lightweight VLM capable of successfully solving online tasks in AndroidWorld. We present a complete pipeline for fine-tuning a pretrained VLM to efficiently tackle goal-based tasks on Android smartphones.
Our results demonstrate that \ourmethod-base achieves the highest AndroidControl action prediction accuracy, compared to all baselines.
Moreover, in online evaluations within AndroidWorld, AppVLM delivers performance comparable to, and in some cases exceeding, agents that rely on GPT-4o, while requiring significantly less time and computational resources. Notably, \ourmethod can compute actions up to ten times faster than GPT-4o agents, making it an efficient alternative for app control.

Despite its strong performance, \ourmethod has limitations stemming primarily from the constraints of its training data. The model struggles with tasks involving operations it has never encountered, such as using the phone's ``clipboard". Since it has not been exposed to the concept of a clipboard, it fails to recognise and execute related actions. Addressing these gaps requires expanding the scope of training data to better capture app control tasks. A promising direction is integrating a broader range of mobile interactions during pretraining, such as UI element detection, UI tree generation, etc.
Existing datasets such as AndroidControl and AitW \citep{aitw} provide valuable benchmarks, but they lack a unified format. For example, AitW does not include UI trees and focuses more on generalisation across Android versions. To advance this field, the community should prioritise the creation of a large-scale, standardised dataset tailored specifically for app control.

Another crucial challenge is data generation. Currently, most datasets rely on human demonstrations, a process that is expensive, time-consuming, and impractical at scale. However, automatically generating trajectories is limited by the lack of reward functions for such tasks. AndroidWorld is the only app control environment that provides an internal reward function. Other approaches leverage large foundation models (e.g., GPT-4o) to evaluate trajectories, but these methods are slow, costly, and highly sensitive to prompt variations, making them unreliable for systematic evaluation.
To overcome these challenges, we believe that the development of dedicated reward models for app control is necessary. Recent studies have explored using models as reward functions \citep{ma2022vip, chan2023vision}, yet no robust, and app control-specific, reward model has been proposed. Such a model would enable scalable evaluation and unlock new possibilities for RL in app control.

\clearpage
\newpage
\newpage

\bibliography{paper}
\bibliographystyle{icml2025}

\newpage
\appendix
\onecolumn
\section{Datasets and Environment}
This section presents additional information and examples about our datasets and environment.

\subsection{Action Space}
\label{appendix:action}
As introduced in \Cref{sec:androidworld_env}, \ourmethod has a fixed action space. This helps standardise actions for training and grounding the model's outputs into valid actions. Our action space is presented in \Cref{tab:action_space}, along with example actions as they are expected to be generated by \ourmethod. The action spaces of AndroidWorld and AndroidControl are very similar, with only minor naming differences, as well as a couple action alterations. AndroidWorld includes a \texttt{keyboard-enter} action which we omit, since it is not present in AndroidControl and thus our initial fine-tuning. AndroidControl also includes a click target as part of its \texttt{input-text} action, while we choose to keep these as separate actions as in AndroidControl.

\begin{table}[h]
\centering
\caption{Action space, along with example actions for each type.} \label{tab:action_space}
\label{tab:model-comparison}
\begin{tabular}{l l}
\toprule
Action & Example\\
\midrule
\openapp + $<$\text{app-name}$>$ & \texttt{\{"action-type":"open-app","app-name":"Clock"\}}\\
\click + $<$\text{target-element}$>$ &\texttt{\{"action-type":"click","target-element":1\}}\\
\longpress + $<$\text{target-element}$>$  &\texttt{\{"action-type":"long-press","target-element":1\}}\\
\inputtext + $<$\text{text}$>$  &\texttt{\{"action-type":"input-text","text":"Hello World"\}}\\
\texttt{scroll-\{up/down/left/right\}}  &\texttt{\{"action-type":"scroll-up"\}}\\
\navigatehome  &\texttt{\{"action-type":"navigate-home"\}}\\
\navigateback  &\texttt{\{"action-type":"navigate-back"\}}\\
\wait  &\texttt{\{"action-type":"wait"\}}\\
\bottomrule
\end{tabular}
\end{table}

The main difference between our action space and that of our evaluation environments lies in our use of click targets for \texttt{click} and \texttt{long-press} actions, rather than x-y coordinates. In such cases, we translate the index of the target element into the centre coordinates of its bounding box, leveraging the provided UI tree information in both AndroidWorld and AndroidControl. Finally, we ensure actions are converted into the specific format expected by either AndroidWorld or AndroidControl, so that actions are correctly executed.

To train on the AndroidControl dataset, it is also necessary to convert the x-y coordinates for ground-truth \texttt{click} and \texttt{long-press} actions to target element indices corresponding to bounding box labels on the screenshot, as expected by our model. The UI element tree information is used to select the best candidate element in this case, and actions which our model can train on are obtained.

\subsection{Observation Space}
\label{appendix:obs}
\begin{figure}[h]
    \centering
    \includegraphics[width=0.8\linewidth]{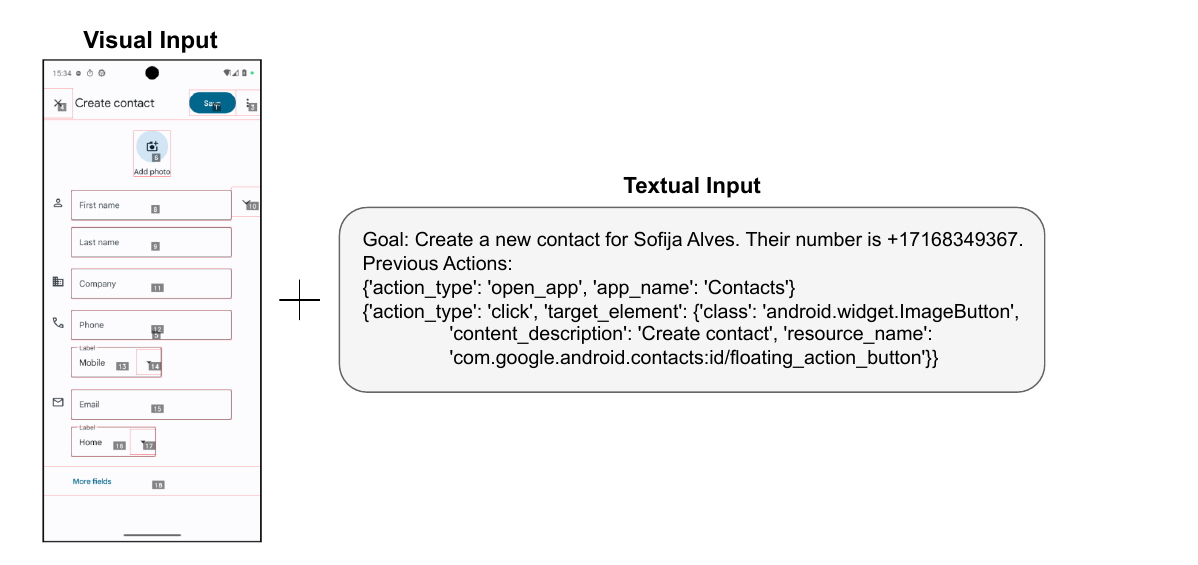}
    \caption{Example AndroidWorld observation passed as input to \ourmethod. The visual input is composed of the current screenshot, annotated with bounding boxes surrounding clickable UI elements, along with numbered labels. The textual input is composed of the task goal, as well as the history of actions. This observation corresponds to the input for step 2 in \Cref{fig:success_contact}.}
    \label{fig:example_obs}
\end{figure}

\Cref{sec:androidworld_env,sec:dataset} introduced the observation processing performed on AndroidWorld and AndroidControl respectively. An example observation from AndroidWorld is illustrated in \Cref{fig:example_obs}, though an observation from AndroidControl would be essentially identical. As previously described, this observation contains both a visual input, the annotated screenshot, and a textual input, composed of the goal and history of actions.

The history of actions provides crucial context for the current state and offers options for error recovery and mitigation. To reduce computational costs, with the objective of creating a lightweight agent, we limit the size of this history to only the five most recent actions. Additionally, the target element index component of \texttt{click} and \texttt{long-press} actions is not very informative as part of this history once the timestep's screenshot is no longer observable. Therefore, the agent stores an alternate representation of actions in its history, sourced from the UI tree data. A condensed textual representation of the target element is used, containing information such as the type of object and its textual content or description, as can be seen in the textual input of \Cref{fig:example_obs}.

\subsection{AndroidWorld Benchmark Set}
\label{appendix:subset}
While the full AndroidWorld benchmark consists of 116 tasks, we use a reduced subset of 82 tasks for our experiments. Firstly, we remove the verification tasks, such as \texttt{ClockStopWatchPausedVerify}, because we check whether tasks have been successfully completed at each timestep and these tasks would automatically succeed. We also remove all Q\&A tasks because they tackle a separate type of task, and are outside of AndroidControl's action space, and thus ours. Finally, we remove the drawing tasks since agents are not equipped for drawing with the current fixed directional scroll actions. The resulting subset of AndroidWorld is used for all agent evaluations, and actually has a higher difficulty distribution than the full set, as shown in \Cref{tab:subset}. As demonstrated by the table, this is because we removed proportionally more ``easy" tasks than ``medium" or ``hard" tasks.

\begin{table}[H]
\centering
\caption{AndroidWorld benchmark count and distribution of tasks per difficulty.}\label{tab:subset}
\begin{tabular}{l cccc }
\toprule
Benchmark & Easy (\%)& Medium (\%)& Hard (\%)& Total\\ 
\midrule
Full Benchmark       & 61 (52.6\%) & 36 (31.0\%) & 19 (16.4\%) & 116\\ 
Our Subset           & 38 (46.3\%) & 28 (34.1\%) & 16 (19.5\%)  &  82\\ 

\bottomrule
\end{tabular}
\end{table}

\section{Implementation Details}

In this section we discuss the implementation details of \ourmethod. As we already discussed we use Paligemma-3b-pt-896\footnote{\url{https://huggingface.co/google/paligemma-3b-pt-896}} as our base model. All fine-tuning rounds, both for the initial SFT, the RFT, and the last SFT steps use the AdamW optimiser \citep{loshchilov2017fixing} with $3\times 10 ^{-6}$ learning rate. The learning rate is gradually reduced to zero during the course of the training. Additionally we fine-tune always for three epochs and we use effective batch size of $64$.  We perform full fine-tuning of the model without using any adapters.

\newpage
\section{Case Studies}
\label{appendix:case_study}
Sample AndroidWorld trajectories from our final model are illustrated in the following figures. \Cref{fig:success_audio_filename} shows our agent correcting the audio recorder task from \Cref{fig:fail_trajectory}. In this example, \ourmethod successfully deletes the existing text before typing the filename. This is particularly impressive because the model has learned to generate a \texttt{long-press} action to do so, an action which is extremely rare in the initial AndroidControl dataset, featuring less than 1\% of the time. It shows the merit of our RFT pipeline, which enables the model to teach itself behaviour it does not have initially. This happens when an agent successfully explores during the data collection phase and the advantageous interaction is reinforced by the rejection sampling and subsequent training.

\vspace{5pt}
\begin{figure*}[ht]
    \centering
    \includegraphics[width=0.98\linewidth]{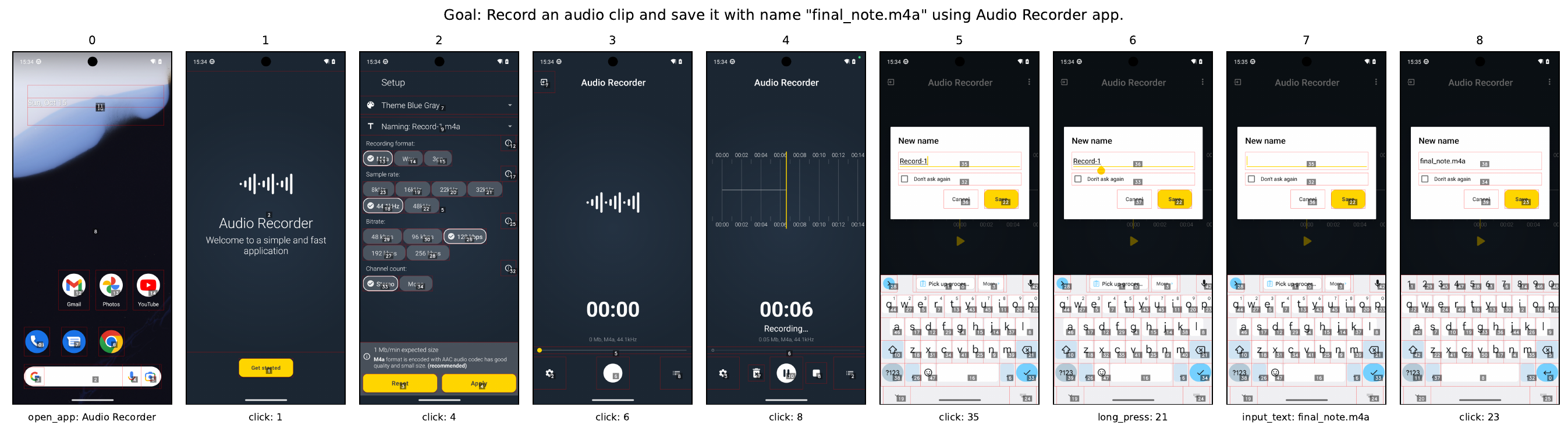}
    \caption{Example trajectory in AndroidWorld, with the goal at the top and the taken actions below each timestep's screenshot. \ourmethod successfully creates an audio recording and saves it with the appropriate filename. Step 6 is noteworthy, with the agent opting for a \texttt{long-press} action, which is very rare in the initial AndroidControl dataset. This figure is in direct juxtaposition with \Cref{fig:fail_trajectory}.}
    \label{fig:success_audio_filename}
\end{figure*}

\vspace{10pt}

\begin{figure*}[ht]
    \centering
    \includegraphics[width=0.9\linewidth]{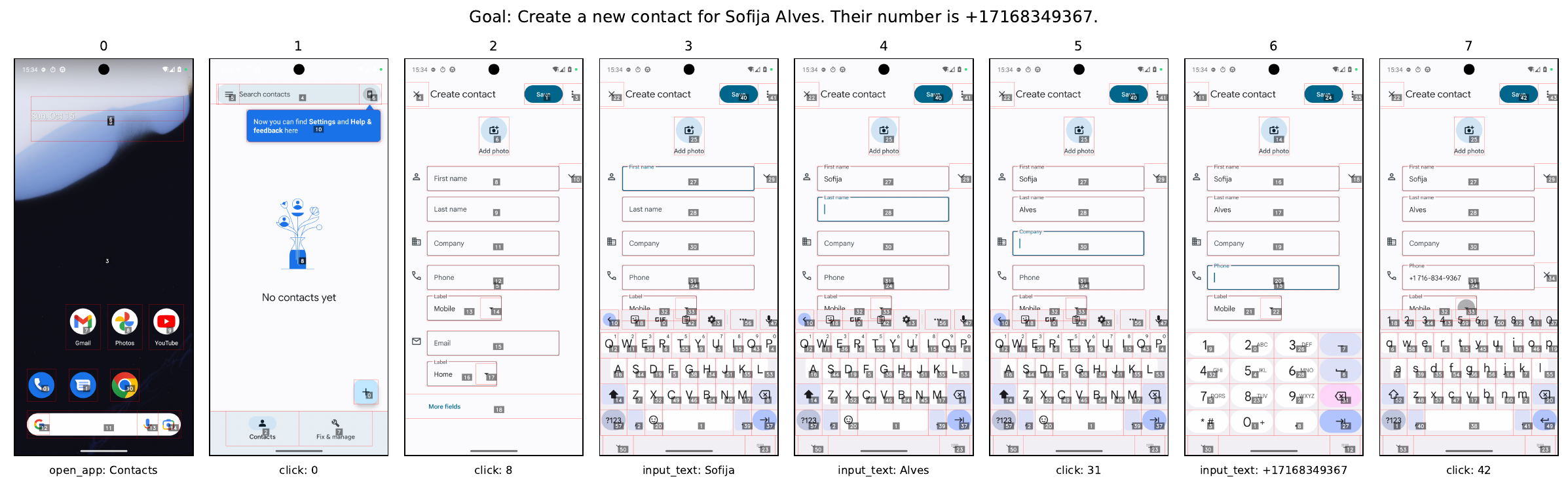}
    \caption{Example trajectory in AndroidWorld, with the goal at the top and the taken actions below each timestep's screenshot. \ourmethod successfully creates a new contact, filling out several form fields to do so. }
    \label{fig:success_contact}
\end{figure*}

\begin{figure*}[ht]
    \centering
    \includegraphics[width=0.9\linewidth]{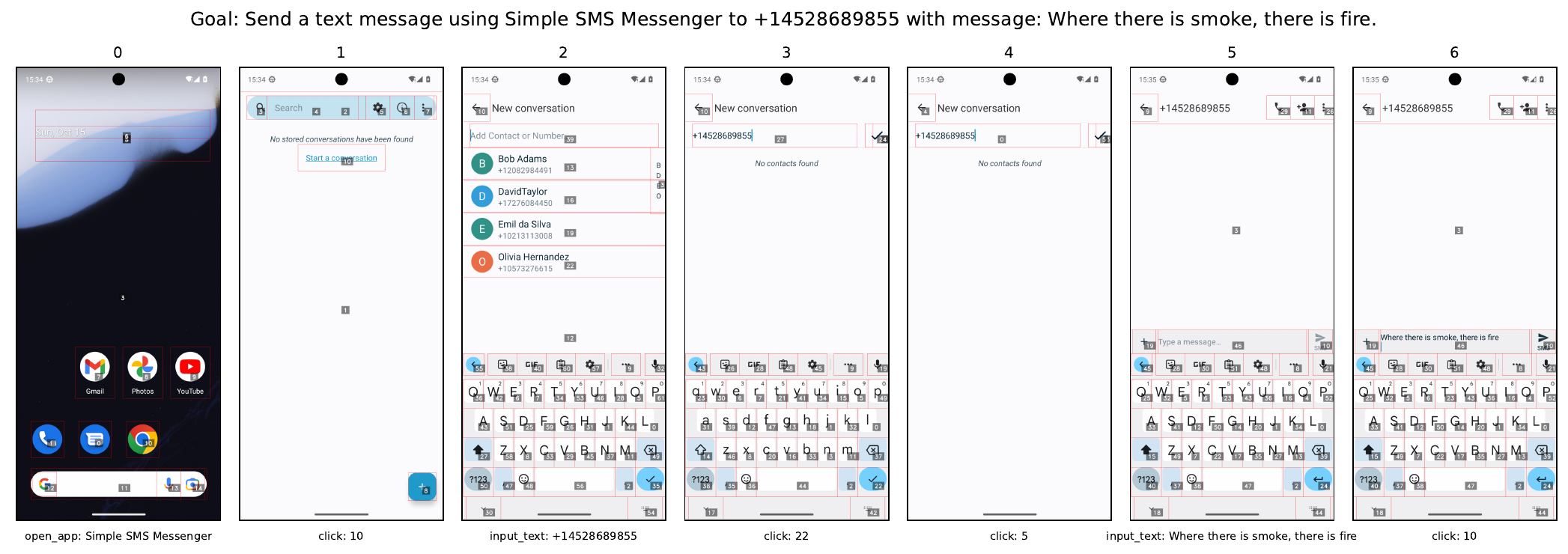}
    \caption{Example trajectory in AndroidWorld, with the goal at the top and the taken actions below each timestep's screenshot. \ourmethod successfully sends a message to a specified phone number.}
    \label{fig:success_sms}
\end{figure*}

\begin{figure*}[ht]
    \centering
    \includegraphics[width=0.9\linewidth]{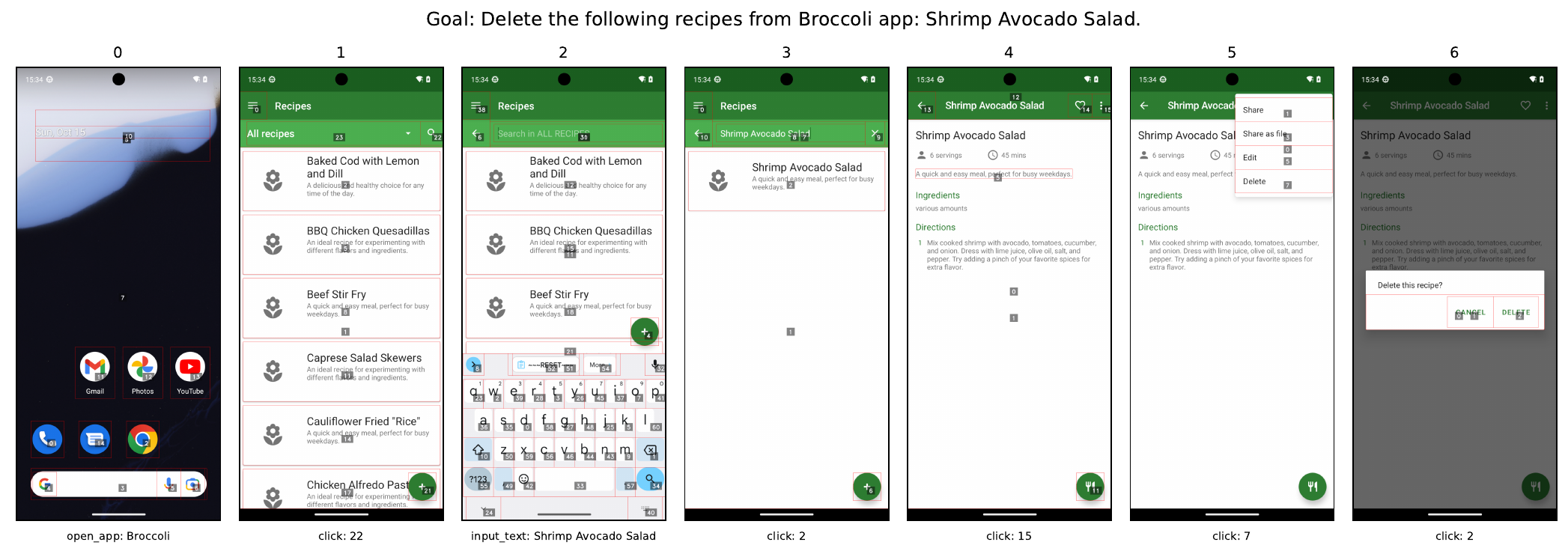}
    \caption{Example trajectory in AndroidWorld, with the goal at the top and the taken actions below each timestep's screenshot. \ourmethod successfully deletes a specific recipe, even when this recipe is not immediately visible in the list}
    \label{fig:success_recipe}
\end{figure*}

Figures \ref{fig:success_contact}-\ref{fig:success_recipe} contain further example trajectories our agent solves successfully in an online fashion. These range from creating a new contact (\Cref{fig:success_contact}), to sending an sms (\Cref{fig:success_sms}) or deleting specific recipes from a dedicated app (\Cref{fig:success_recipe}).

\vspace{10pt}


\end{document}